\documentclass[conference]{IEEEtran}
\IEEEoverridecommandlockouts

\usepackage{cite}
\usepackage{amsmath,amssymb,amsfonts}
\usepackage{algorithmic}
\usepackage{graphicx}
\usepackage{textcomp}
\usepackage{xcolor}
\usepackage{texnames}
\usepackage{siunitx}
\usepackage{booktabs}
\usepackage{flushend}
\usepackage{multirow}
\usepackage[colorlinks=true, urlcolor=blue, linkcolor=black]{hyperref}

\begin{document}

\title{	Physics-Informed Diffusion Models for SAR Ship Wake Generation from Text Prompts
\thanks{
    Kamirul Kamirul, Odysseas Pappas, and Alin Achim are with the
    Visual Information Laboratory, University of Bristol, BS1 5DD Bristol,
    U.K. (e-mail: kamirul.kamirul@bristol.ac.uk; o.pappas@bristol.ac.uk;
    alin.achim@bristol.ac.uk). (Corresponding author: Kamirul Kamirul.)
}
\thanks{
Kamirul Kamirul is fully funded by Lembaga Pengelola Dana Pendidikan (LPDP), Ministry of Finance of Republic of Indonesia.
}
}


\author{\IEEEauthorblockN{Kamirul Kamirul, Odysseas Pappas, Alin Achim}
\IEEEauthorblockA{\textit{Visual Information Laboratory} \\
\textit{University of Bristol}, UK \\
(kamirul.kamirul, o.pappas, alin.achim)@bristol.ac.uk}
}

\maketitle

\begin{abstract}

Detecting ship presence via wake signatures in SAR imagery is attracting considerable research interest, but limited annotated data availability poses significant challenges for supervised learning.
Physics-based simulations are commonly used to address this data scarcity, although they are slow and constrain end-to-end learning.
In this work, we explore a new direction for more efficient and end-to-end SAR ship wake simulation using a diffusion model trained on data generated by a physics-based simulator.
The training dataset is built by pairing images produced by the simulator with text prompts derived from simulation parameters.
Experimental result show that the model generates realistic Kelvin wake patterns and achieves significantly faster inference than the physics-based simulator. 
These results highlight the potential of diffusion models for fast and controllable wake image generation, opening new possibilities for end-to-end downstream tasks in maritime SAR analysis.

\end{abstract}

\begin{IEEEkeywords}
diffusion models, physics-based simulation, ship wake, synthetic
aperture radar, text-to-image.
\end{IEEEkeywords}

\section{Introduction}

Wakes can serve as an indirect cue of ship presence in Synthetic Aperture Radar (SAR) imagery, which supports maritime security applications such as vessel monitoring, classification, and anomaly detection~\cite{SynthwakeSAR_Paper}. 
Unfortunately, the development of robust ship wake detection algorithms is constrained by the limited availability of annotated SAR datasets. 

To address this, physics-based simulators (PBS) have been used to provide synthetic images by modeling the interaction between a moving vessel and the sea surface~\cite{tunaley_1991, shemer1996, hennings1999}. 
While PBS can produce accurate and realistic wake patterns, they are computationally expensive and require parameter tuning. 
The high computational cost of the simulators stems from the sea wave modeling process which involves summing many independent harmonic waves using random phase modeling. 
Additionally, since manual tuning of the parameters is required, these simulators are not well suited for pipelines requiring end-to-end integration.

Motivated by the challenges associated with physics-based simulators, we explore the use of diffusion models as an alternative approach for simulating SAR ship wakes.
Our contribution lies in demonstrating that existing diffusion frameworks can be adapted for controllable and efficient ship wake image generation.
To this end, we employed Stable Diffusion (SD) v1.4, a state-of-the-art Latent Diffusion Model (LDM)\cite{Stable_Diffusion}, as our diffusion backbone.
The choice of latent-based diffusion over image-based diffusion is motivated by its faster inference time and well-known capability for text-to-image generation, aligning with our goal of text-driven simulation.
We create a dataset of image-text pairs by generating wake images with PBS and pairing them with descriptive text prompts derived from the generation parameters.
The dataset is ultimately used to train the LDM to learn Kelvin wake patterns.

Experimental results demonstrate that the LDM achieves a significant speedup over the PBS while maintaining competitive visual quality in wake image generation. 
This capability provides scalable wake synthesis while eliminating the computational overhead associated with physics-based methods.


\section{Theoretical Preliminaries}

\subsection{Physics-based Wake Simulation}

Physics-based approaches combine hydrodynamic modeling of surface wave generation with electromagnetic backscatter mechanisms to replicate key features such as Kelvin wakes and turbulent trails.

Early simulation efforts focused on simplified imaging scenarios. Tunaley \textit{et al.}~\cite{tunaley_1991} modeled sea clutter and wake interactions, while Shemer \textit{et al.}~\cite{shemer1996} introduced interferometric SAR (InSAR) techniques for wake detection. These approaches were later validated by Hennings\textit{ et al.}~\cite{hennings1999}, who compared simulated Kelvin wakes with observed SAR data across multiple platforms.


A significant advancement in wake simulation involves the two-scale model (TSM) approach, which separates large-scale deterministic wave components from small-scale stochastic roughness (cf.~\cite{RIZAEV2022120} and references therein). 
Modern simulation frameworks now combine TSM with linear sea surface theory and Michell’s thin-ship wake model, allowing for flexible parameterization of SAR imaging conditions (e.g., frequency, polarization, incidence angle) and ocean state variables (e.g., wind speed, Froude number). 
These models further incorporate SAR-specific effects, including velocity bunching and azimuthal cut-off, which affect the resolution and displacement of moving wave features\cite{RIZAEV2022120, SynthwakeSAR_Paper, Kamirul_SAR_Ship_S_Band}.


\subsection{Diffusion Models}

Diffusion models have emerged as a powerful class of generative models for high-quality image synthesis, achieving state-of-the-art results across various domains. 
These models learn to reverse a fixed forward noising process that gradually transforms data into Gaussian noise~\cite{DDPM}. 
During inference, the model reconstructs data from noise through a learned denoising process.

Given a clean sample \( \mathbf{x}_0 \), the forward process adds Gaussian noise over \( T \) steps:
\begin{equation}
q(\mathbf{x}_t | \mathbf{x}_{t-1}) = \mathcal{N}(\mathbf{x}_t; \sqrt{1 - \beta_t} \mathbf{x}_{t-1}, \beta_t \mathbf{I}),
\end{equation}
with a closed-form expression:
\begin{equation}
\mathbf{x}_t = \sqrt{\bar{\alpha}_t} \mathbf{x}_0 + \sqrt{1 - \bar{\alpha}_t} \boldsymbol{\epsilon}, \quad \boldsymbol{\epsilon} \sim \mathcal{N}(0, \mathbf{I}),
\end{equation}
where \( \bar{\alpha}_t = \prod_{s=1}^{t}(1 - \beta_s) \).

The model \( \epsilon_\theta \) is trained to predict the added noise using the simplified loss:
\begin{equation}
\mathcal{L} = \mathbb{E}_{\mathbf{x}_0, t, \boldsymbol{\epsilon}} \left[ \left\| \boldsymbol{\epsilon} - \epsilon_\theta(\mathbf{x}_t, t) \right\|^2 \right].
\end{equation}
The network \( \epsilon_\theta \) is typically implemented as a U-Net, which enables multi-scale feature learning and effective denoising across spatial resolutions.
For conditional generation, such as text-to-image synthesis, conditioning information \( \mathbf{c} \) (e.g., a text embedding) is incorporated:
\begin{equation}
\epsilon_\theta(\mathbf{x}_t, t, \mathbf{c}).
\end{equation}

While early diffusion models operate in pixel space, Latent Diffusion Models (LDMs)~\cite{Stable_Diffusion} perform denoising in a learned latent space.
A pre-trained autoencoder encodes images into low-dimensional latent vectors, enabling diffusion to operate in a compact domain.
This reduces memory and computation costs significantly during both training and inference, while preserving high-fidelity outputs.


\section{Methodology}

\begin{figure}[htb]
\centering
\includegraphics[width=0.47\textwidth, trim={0 0 0 0}]{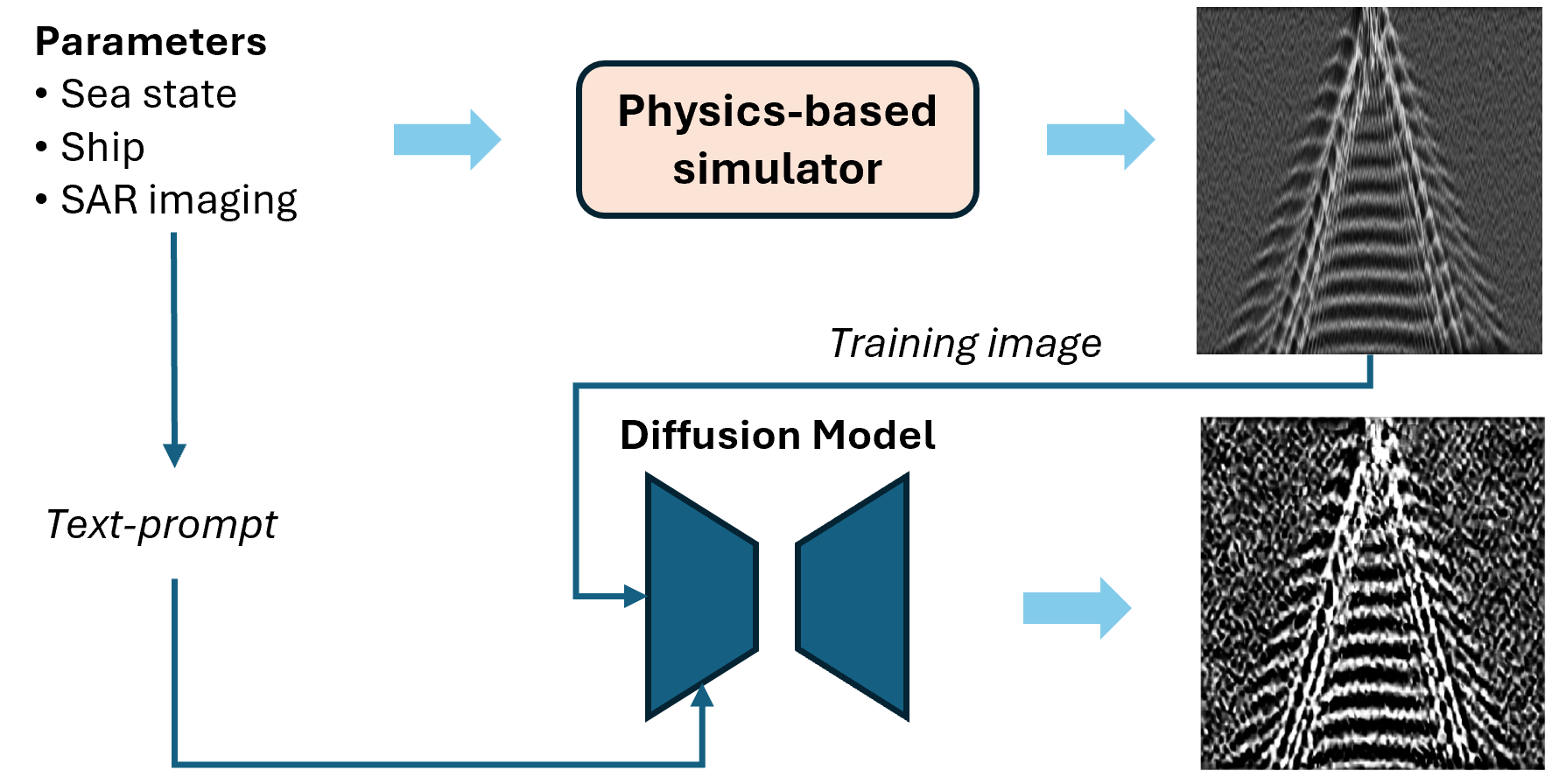}
\caption{
   Training of a text-conditioned diffusion model for SAR ship wake generation. Synthetic images from a physics-based simulator enable a physics-informed learning process.
}
\label{fig:flowchart_PBD_LDM}
\end{figure}

The overall process for model training and evaluation is given in Fig. \ref{fig:flowchart_PBD_LDM}.
We begin with construction of training dataset utilizing existing simulator and then train the model using the images and textual input derived from simulation parameters.

\subsection{Training images}

This work employs the SAR ship wake simulator developed by Rizaev and Achim~\cite{AssenSAR_Code}.
The simulator models wakes as a superposition of a frozen wind-driven sea surface and a vessel-generated Kelvin wave pattern~\cite{RIZAEV2022120}.
SAR images are synthesized using the Two-Scale Model (TSM), incorporating resonant Bragg scattering and accounting for tilt and hydrodynamic modulations.
The simulator takes sea state, ship, and SAR imaging parameters as inputs. In this work, we use the configuration detailed in Table~\ref{table:sim_params} to generate 3,450 synthetic SAR images for model training.

\begin{table}[htbp]
\caption{Parameters Used in Generating Synthetic Images}
\label{table:sim_params}
\centering
\begin{tabular}{l l}
\hline
\textbf{Parameter} & \textbf{Values} \\
\hline
Ship length & 195 m \\
Ship breadth & 26 m \\
Ship speed & 6, 8, 10, 12 m/s \\
Ship heading & 0$^\circ$, 45$^\circ$, 90$^\circ$, 135$^\circ$, 180$^\circ$, 225$^\circ$, 270$^\circ$, 315$^\circ$ \\
Wind speed & 3, 5, 7 (m/s) \\
Wind direction & 0$^\circ$, 90$^\circ$, 180$^\circ$, 270$^\circ$ \\
Fetch length & 10,000 m, 40,000 m, 70,000 m \\
Frequency & 9.6 GHz \\
Polarization & VV \\
Incident angle & 16$^\circ$, 24$^\circ$, 32$^\circ$ \\
\hline
\end{tabular}
\end{table}

\subsection{C. Parameter-to-Text Conversion}

For each simulated image, the corresponding simulation parameters are extracted and converted into a structured natural language description. These prompts are formatted to comply with the 77-token maximum input length of the CLIP text encoder. An example prompt is:

\textit{``Simulated SAR image of wakes from a ship moving at 10 mps toward 135 degrees. Ocean waves are shaped by 3 mps wind directing toward 90 degrees and 10000 meter fetch. Captured at 32 degree incidence.''}


\subsection{Model Training}


The model is trained for 4000 epochs with a batch size of 16 and a learning rate of \(1 \times 10^{-5}\), using PyTorch 2.4 on Rocky Linux 8.9. Training takes place on dual NVIDIA V100 GPUs hosted by the University of Bristol’s Advanced Computing Research Centre HPC cluster. Furthermore, performance benchmarking is conducted on a single NVIDIA RTX 3090 GPU.

\section{Results and Discussion}

This section evaluates the Latent Diffusion Model (LDM) by comparing its image quality and inference speed against the physics-based simulator (PBS), followed by a discussion of the results.

\subsection{Visual Quality}

\begin{figure}[ht]
    \centering
    \begin{tabular}{@{}p{.25\linewidth}@{\hspace{0.5em}}p{.25\linewidth}@{\hspace{0.5em}}p{.25\linewidth}@{}}
        \includegraphics[width=\linewidth]{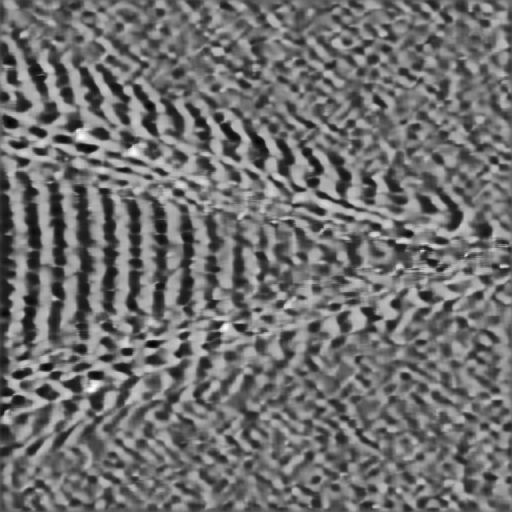} &
        \includegraphics[width=\linewidth]{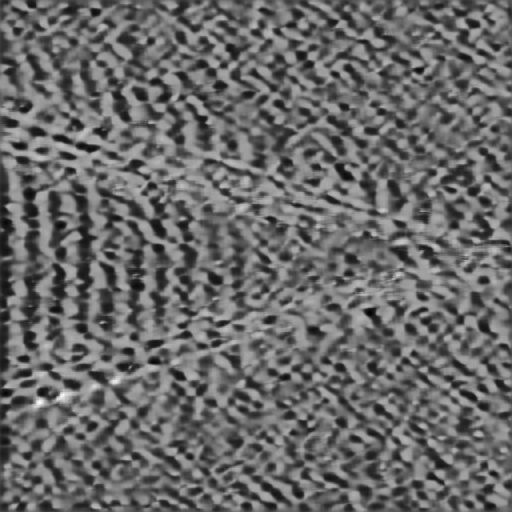} &
        \includegraphics[width=\linewidth]{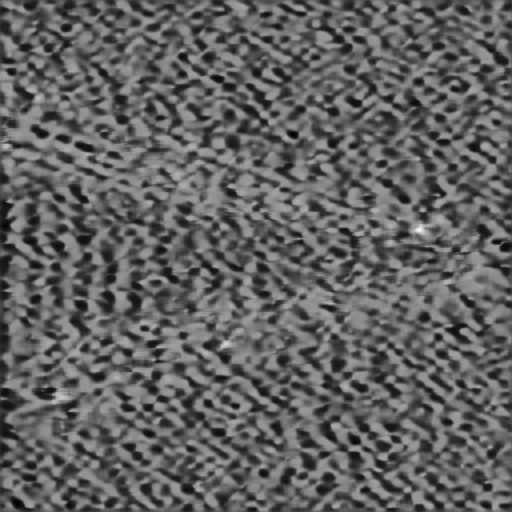} \\
        \centering\scriptsize $v_w = 3\, \mathrm{m/s}$ & \centering\scriptsize $v_w = 5\, \mathrm{m/s}$ & \centering\scriptsize $v_w = 7\, \mathrm{m/s}$
    \end{tabular}

    \vspace{1ex}  

    \begin{tabular}{@{}p{.25\linewidth}@{\hspace{0.5em}}p{.25\linewidth}@{\hspace{0.5em}}p{.25\linewidth}@{}}
        \includegraphics[width=\linewidth]{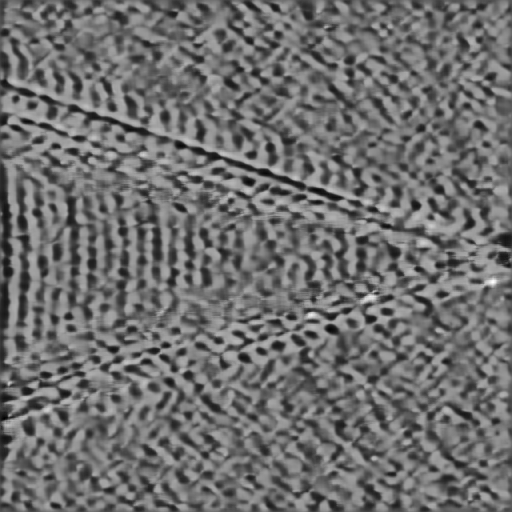} &
        \includegraphics[width=\linewidth]{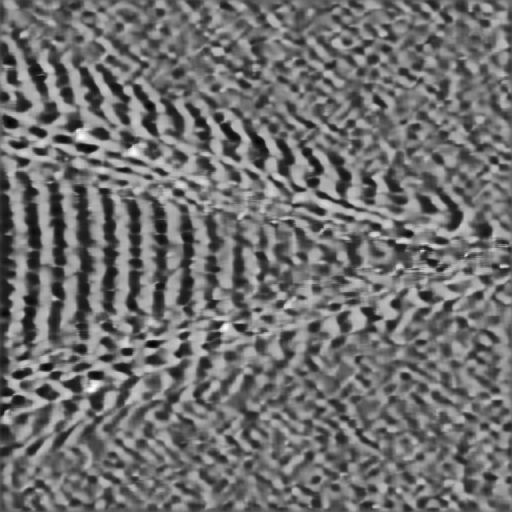} &
        \includegraphics[width=\linewidth]{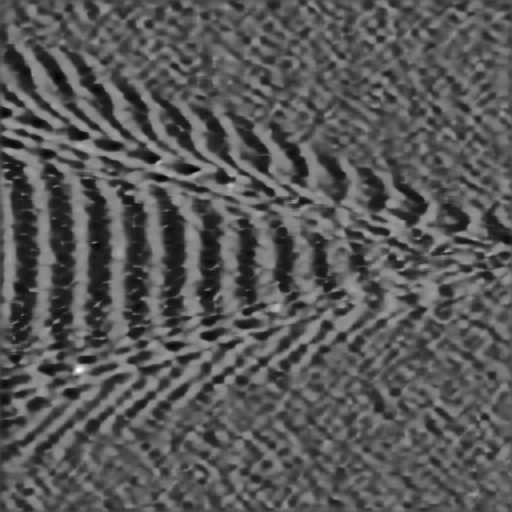} \\
        \centering\scriptsize $v_s = 8\, \mathrm{m/s}$ & \centering\scriptsize $v_s = 10\, \mathrm{m/s}$ & \centering\scriptsize $v_s = 12\, \mathrm{m/s}$
    \end{tabular}

    \vspace{1ex}  

    \begin{tabular}{@{}p{.25\linewidth}@{\hspace{0.5em}}p{.25\linewidth}@{\hspace{0.5em}}p{.25\linewidth}@{}}
        \includegraphics[width=\linewidth]{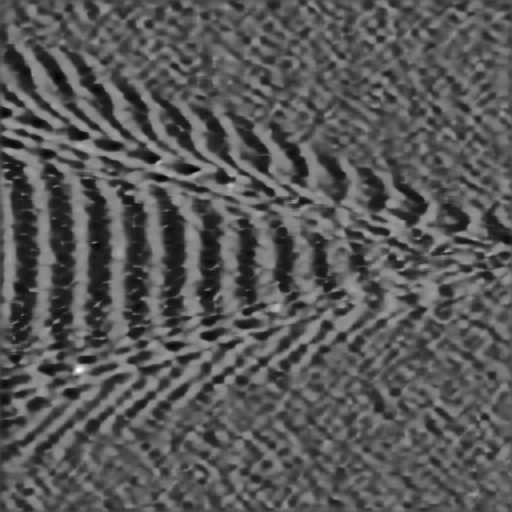} &
        \includegraphics[width=\linewidth]{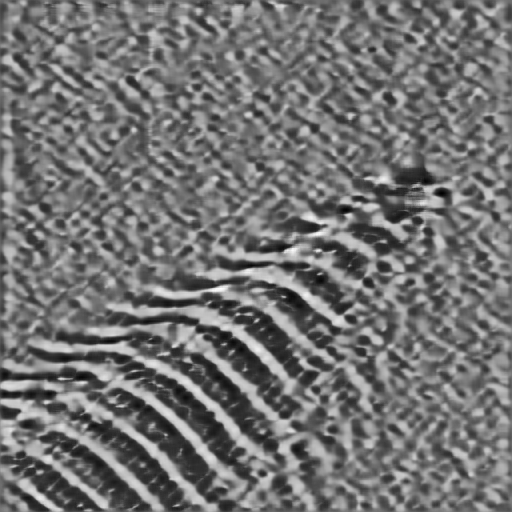} &
        \includegraphics[width=\linewidth]{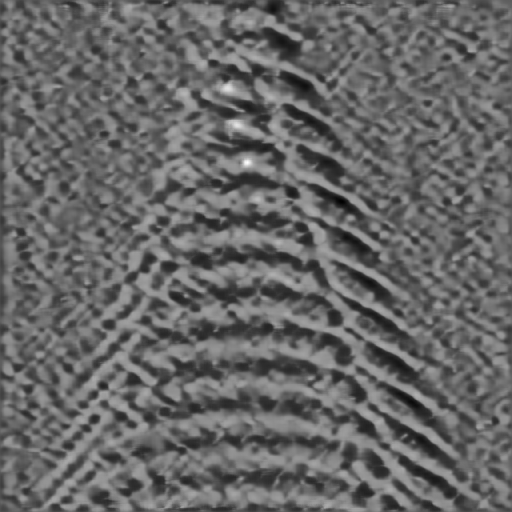} \\
        \centering\scriptsize $\theta_s = 0^\circ$ & \centering\scriptsize $\theta_s = 45^\circ$ & \centering\scriptsize $\theta_s = 90^\circ$
    \end{tabular}

    \vspace{0.0ex}  
\caption{\scriptsize
Effect of sea state and ship parameters on wake patterns generated by the LDM. 
\textbf{Top row:} Variation of wind speed ($v_w = 3$~m/s, $5$~m/s, $7$~m/s), illustrating how increasing wind strength influences the surface wave characteristics. 
\textbf{Middle row:} Variation of ship speed ($v_s = 8$~m/s, $10$~m/s, $12$~m/s), showing how changes in vessel velocity affect the wake structure. 
\textbf{Bottom row:} Variation of ship orientation ($\theta_s = 0^\circ$, $45^\circ$, $90^\circ$), demonstrating the impact of heading angle on wake symmetry and direction.
}
\label{fig:param_effect}
\end{figure}

Fig.~\ref{fig:param_effect} illustrates example outputs generated by the LDM under variations of selected input parameters. 
The generated wake patterns demonstrate visual consistency with the conditioning inputs, suggesting that the model has learned meaningful structural representations characteristic of SAR ship wake imagery. 
Specifically, increased wind speeds result in visibly rougher surface textures, higher ship speeds lead to increasingly visible Kelvin structures, and the orientation of the wake aligns with the specified vessel heading. 
These observations indicate that LDM exhibits consistent behavior with the physical dependencies encoded in the training data.

\begin{figure}[ht]
    \centering
    \begin{tabular}{@{}p{.33\linewidth}@{\hspace{0.5em}}p{.33\linewidth}@{}}
        \includegraphics[width=\linewidth]{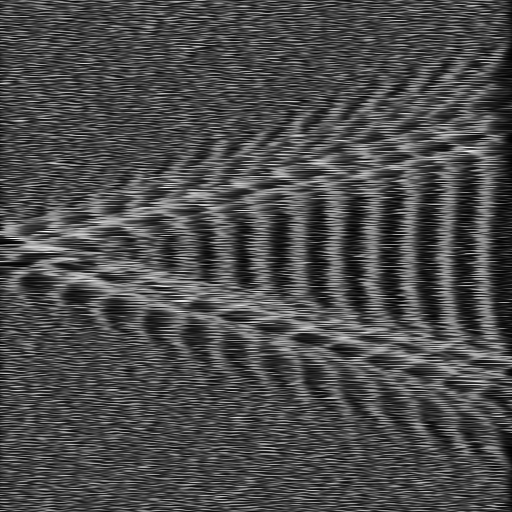} &
        \includegraphics[width=\linewidth]{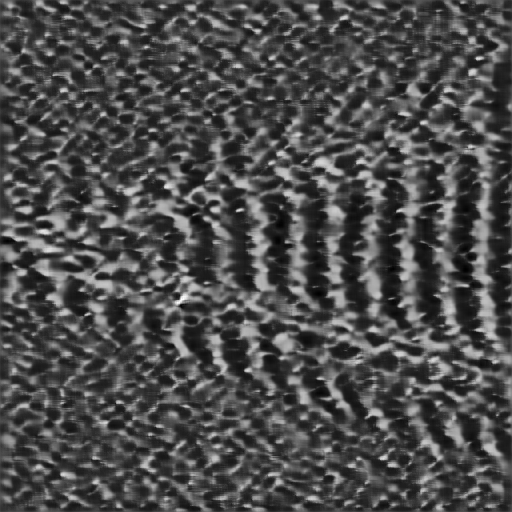} \\
    \end{tabular}

    \vspace{1ex}  
        \begin{tabular}{@{}p{.33\linewidth}@{\hspace{0.5em}}p{.33\linewidth}@{}}
        \includegraphics[width=\linewidth]{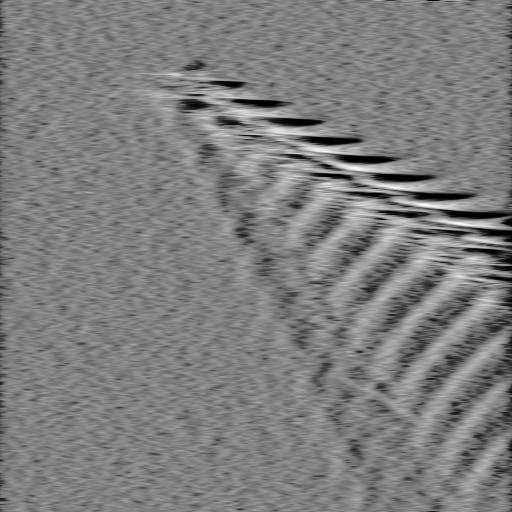} &
        \includegraphics[width=\linewidth]{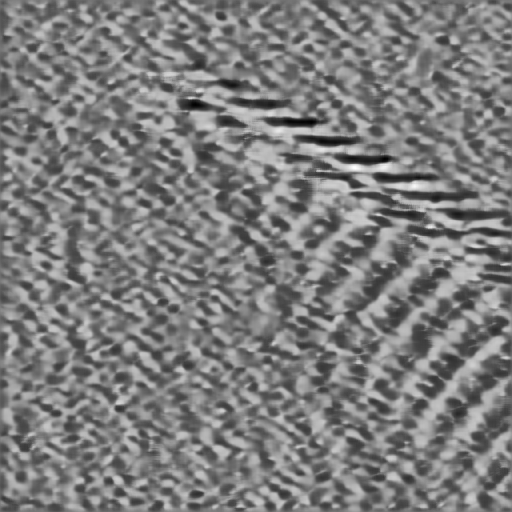} \\
    \end{tabular}

    \vspace{0ex}  
\caption{\scriptsize
Comparison of SAR wake simulations for two prompt conditions, with PBM results on the left and LDM results on the right in each row.}
\label{fig:traditional_vs_diffusion_comparison}
\end{figure}

Furthermore, Fig.~\ref{fig:traditional_vs_diffusion_comparison} compares the images from both methods under the same input parameters.
The LDM outputs (left column) show strong structural similarity to the PBS outputs, accurately capturing features such as the Kelvin wake shape, orientation, and intensity distribution.
This strong visual alignment demonstrates that the model has successfully internalized the spatial and statistical patterns encoded in the physics-based simulations.
Crucially, this resemblance highlights the model’s ability to generate realistic SAR wake patterns guided indirectly by physical principles—learned through exposure to simulator-generated training data. 

\subsection{Quantitative Evaluation}


To assess the structural accuracy and perceptual quality of images generated by LDM, we employed SSIM, PSNR and LPIPS metrics and the results is presented in Table~\ref{tab:metrics}.
\begin{table}[ht]
\centering
\caption{Image Quality Metrics}
\label{tab:metrics}
\begin{tabular}{lccc}
\toprule
Metric & Score \\
\midrule
SSIM & 0.17 \\ 
PSNR (dB) & 12.57 \\ 
LPIPS & 0.57 \\ 
\bottomrule
\end{tabular}
\end{table}
The SSIM score of 0.17, PSNR of 12.57 dB, and LPIPS of 0.57 all indicate substantial differences in structure and perceptual quality. 
These results are consistent with the visual comparison shown in Fig.~\ref{fig:traditional_vs_diffusion_comparison}, where the reconstructed images appear noticeably blurry, particularly in fine-scale regions.
This degradation in quality can be attributed to the limited scope of the current training, as the model was trained with a relatively small dataset and over a modest number of epochs. 
Given that the purpose of this experiment is to demonstrate the basic feasibility of applying diffusion models to ship wake simulation, these initial results are encouraging. 
It is expected that increasing the dataset size and extending the training duration would significantly enhance the model’s ability to preserve sharper structures and finer details, reducing blurriness and improving both structural and perceptual fidelity.

\begin{figure}[htb]
\centering
\includegraphics[width=0.45\textwidth, trim={0 0 0 0}]{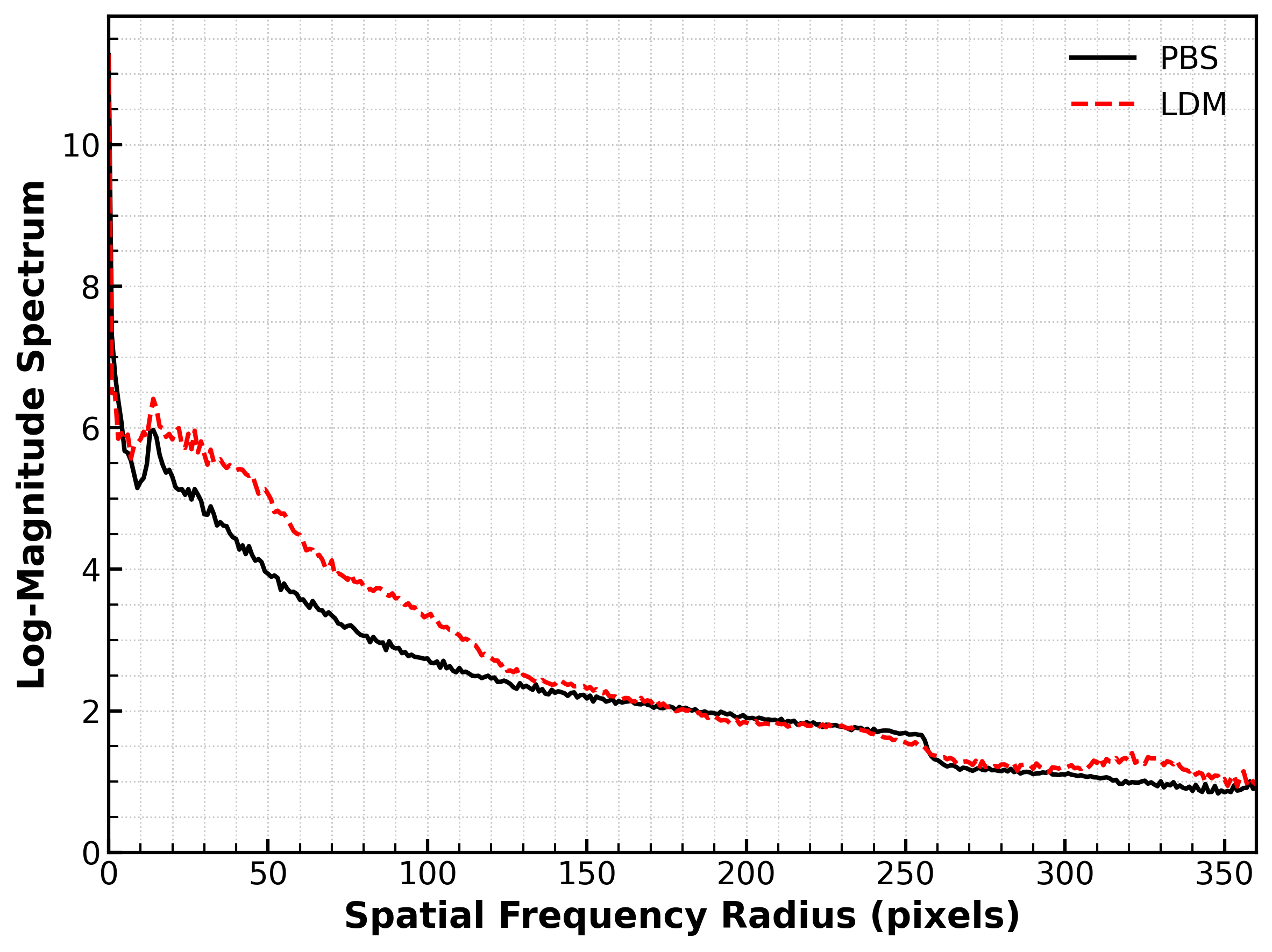}
\caption{
    Spectral comparison between images generated by physics-based simulator (PBS) and latent diffusion model (LDM).
}
\label{fig:radial_spectrum_comparison}
\end{figure}

Spectral analysis on the images generated by the two methods provides further insight into LDM capability, as shown in Fig.~\ref{fig:radial_spectrum_comparison}.
This analysis is performed by applying a two-dimensional Fourier transform and computing the radial average of the magnitude spectrum as a function of spatial frequency. This procedure follows standard practice in frequency domain analysis~\cite{ZHANG20101186,Taylor_NCBI}.

At very low spatial frequencies (0–10 pixels), the LDM and PBS spectra align closely, indicating that the LDM successfully reproduces the largest-scale wake structures, such as the overall shape and direction of the ship wakes.

However, in the mid-frequency range (10–150 pixels), the LDM-generated image exhibit elevated spectral energy compared to that of PBS. This suggests that the LDM introduces excess medium-scale noise, likely due to an imperfect denoising process, as visually confirmed in Fig.~\ref{fig:traditional_vs_diffusion_comparison}.

At higher frequencies (radius 150–360), the spectral profiles between the two converge more closely. 
This convergence implies that while the model does not fully eliminate high-frequency noise, it maintains a comparable level of fine structural details without significant overshooting.

Overall, the conducted spectral analysis confirms that the LDM successfully captures large-scale wake structures; however, improvement in the training process are required to reduce noise in the generated images.

\subsection{Inference Speed}

Table~\ref{tab:speed} shows that LDM is consistently faster than PBS, with performance gains becoming more pronounced as image size increases.
In PBS, the sea surface is explicitly synthesized by summing a large number of independent harmonic components, following two-scale model theory, where gravity and capillary waves are separately modeled and superimposed. 
This summation introduces high computational complexity, scaling with both the number of wave components and the image resolution.
In contrast, LDM avoids explicit wave synthesis by directly generating wake patterns through a fast, learned denoising process in latent space.
As a result, LDM achieves up to an 11× speedup over PBS for $512\times 512$ image sizes.

\begin{table}[ht]
\centering
\caption{Average Inference Time Per Image}
\label{tab:speed}
\begin{tabular}{c c c c}
\toprule
\multirow{2}{*}{\textbf{Image Size}} & \multicolumn{2}{c}{\textbf{Speed (s)}} & \multirow{2}{*}{\textbf{Speed gain}} \\ 
\cmidrule(lr){2-3}
 & \textbf{LDM} & \textbf{PBM} & \\ 
\midrule
128 $\times$ 128 & 3.37 & 1.12 & 3.01 \\
256 $\times$ 256 & 8.81 & 1.49 & 5.91 \\
512 $\times$ 512 & 28.47 & 2.59 & 10.99 \\
\bottomrule
\end{tabular}
\end{table}

Though primarily exploratory, this work reveals several promising directions. 
First, experimental results demonstrate that LDM enables controllable wake generation through text prompts, providing a flexible and scalable tool for creating labeled SAR wake imagery. 
This capability is particularly valuable in scenarios where real data is limited, such as rare vessel types or extreme environmental conditions.

Second, LDM achieves inference speeds nearly 11 times faster than PBS, making it a highly viable alternative for efficient SAR ship wake generation. 
The combination of faster inference and end-to-end compatibility, absent in PBS, makes LDM a more practical tool for modern simulation pipelines.

Lastly, this study shows that the limitations of PBS---particularly its slow inference and lack of end-to-end compatibility---can be addressed by transferring its knowledge into a diffusion model, as demonstrated in this work. 
This leads to faster, accurate, and fully integrated wake simulator, achieving a win-win strategy that preserves physical accuracy while enabling scalable and efficient pipelines, which are difficult to realize with physics-based methods alone.






\section{Conclusions}
This work explored the use of text-guided latent diffusion models (LDMs) as a data-driven alternative for simulating SAR ship wakes. 
Trained on images generated by a physics-based simulator (PBS), the model learns to synthesize wake patterns conditioned on text prompts derived from simulation parameters. 
Results show that the LDM results in comparable images while offering significantly faster inference compared to PBS.
These findings suggest that LDMs can serve as a practical alternative to traditional physics-based simulators, particularly when fast generation and end-to-end integration are required—capabilities that conventional methods struggle to support. While physics-based simulation remains a critical foundation, encoding its knowledge into a diffusion model, as demonstrated in this work, provides a path toward faster and end-to-end-compatible SAR wake generation.


\bibliographystyle{IEEEtranS}
\bibliography{MyReferences}

\end{document}